\documentclass[a4paper,11pt]{article}

\usepackage{fourier}
\usepackage{color}
 \usepackage{graphicx}
\usepackage{url}
\usepackage[affil-it]{authblk}
\usepackage{amsmath}
\usepackage{wrapfig}
\usepackage{xspace}

\usepackage[T1]{fontenc}
\usepackage{times}

\begin{document}

\title{Entropic Regularisation of Robust Optimal Transport}

%\author{Anonymous Submission}
 \author{Rozenn Dahyot, Hana Alghamdi and Mairead Grogan}
%\affil{Anonymous Affiliation}
\affil{School of Computer Science and Statistics \\
Trinity College Dublin, Ireland\\
 Rozenn.Dahyot@tcd.ie, alghamdh@tcd.ie,  groganma@tcd.ie}
\date{}
\maketitle
\thispagestyle{empty}

\begin{abstract}
 Grogan et al. \cite{GROGAN201939,Grogan:2017:UII:3150165.3150171} have recently proposed a solution to colour transfer by minimising the Euclidean distance $\mathcal{L}_2$ between two probability density functions capturing the colour distributions of two images (palette and target). 
 It was shown to be very competitive to alternative solutions based on Optimal Transport for colour transfer. 
 We show that in fact Grogan et al's formulation can also  be understood as a new robust Optimal Transport based framework with  entropy regularisation over marginals.
\end{abstract}

%%%%%%%%%%%%%%%%%%%%%%
\section{Introduction}
\begin{wrapfigure}{r}{0.5\textwidth}
  \vspace{-20pt}
  \begin{center}
    \includegraphics[width=0.4\textwidth]{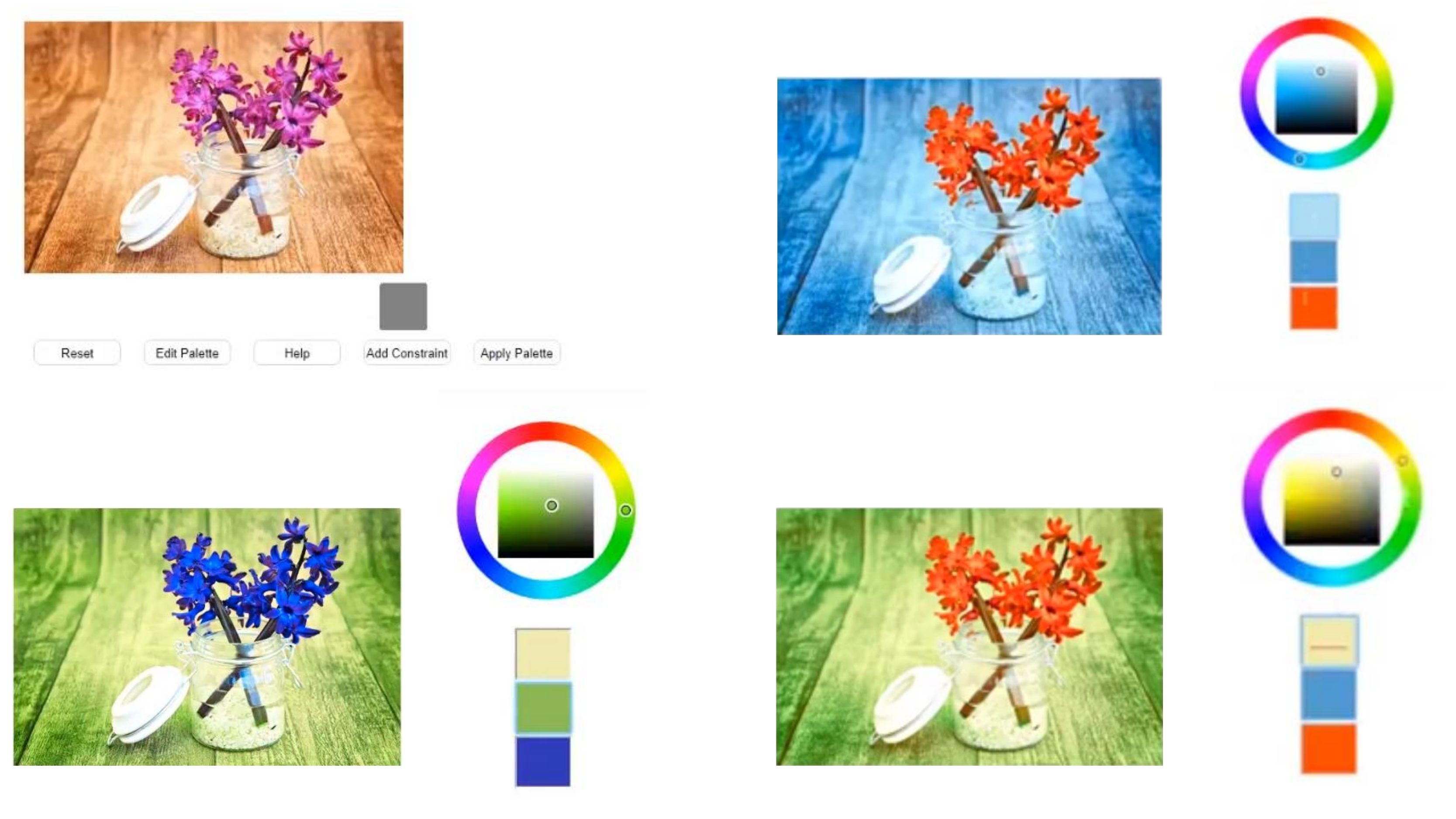}
\end{center}
\vspace{-20pt}
  \caption{\small{Colour correspondences with  colour palettes allows user interactions when recolouring   a reference image (top left) \cite{Grogan:2017:UII:3150165.3150171}.}}\label{fig:demo}
  \vspace{-0pt}
\end{wrapfigure}
%%%%%%%%%%%%%%%%%%%%%%

Optimal transport (OT) \cite{MAL-073} has  been successfully used  as a way for defining cost functions for optimisation when performing colour transfer \cite{PITIE2007123} and more recently in machine learning \cite{7586038,MAL-073}.  
The optimal transport cost (e.g Wasserstein distance) itself is also used  as a similarity  metric for retrieval  \cite{Rubner:2000:EMD:365875.365881}.
For colour transfer (see Fig. \ref{fig:demo}\footnote{Images extracted from  video \url{https://youtu.be/FfrdyKMBVRc} (demo for \cite{Grogan:2017:UII:3150165.3150171}) have been used for designing Fig. \ref{fig:demo}.}), Grogan et al.
\cite{GROGAN201939,Grogan:2017:UII:3150165.3150171}
have recently proposed an alternative approach for designing the cost  function based on the $\mathcal{L}_2$ divergence (see section \ref{sec:L2}). 
This $\mathcal{L}_2$ based cost function is 
a weighted sum of multiple terms (terms $\mathcal{T}_{0,1,2,3}$ in Eq. \ref{eq:1}) able to take into account correspondences between images (via term $\mathcal{T}_3$ in Eq. \ref{eq:1}) when these are available, as well as the unsupervised scenario when no correspondence is available (via term $\mathcal{T}_2$ in Eq. \ref{eq:1}). In addition, $\mathcal{L}_2$  includes  entropies   (terms $\mathcal{T}_0$  and $\mathcal{T}_1$ in Eq. \ref{eq:1}). To further constrain the cost function when estimating the colour transformation $\phi$,   additional penalties can be added to prevent colours exceeding a certain range or forcing the estimated solution $\phi$ to be smooth (resp. terms $\mathcal{T}_4$ and $\mathcal{T}_5$ in Eq. \ref{eq:1}).  The estimate $\hat{\phi}$ is computed as $$\hat{\phi}=\arg \min_{\phi}\mathcal{C}(\phi) $$
with: 
\begin{equation}\label{eq:1}
\begin{array}{l|lcl}
\mathcal{C}(\phi)= 
& 
  +\frac{1}{n^2} \sum_{j_1=1}^{n} \sum_{j_2=1}^{n} \mathcal{N}(0;y^{(j_1)}-y^{(j_2)},2h^2 \mathrm{I})  & &(\mathcal{T}_0)\\ 
&&&\\  
& 
  +\frac{1}{\tilde{n}^2} \sum_{i_1=1}^{\tilde{n}} \sum_{i_2=1}^{\tilde{n}} \mathcal{N}(0;\tilde{y}^{(i_1)}-\tilde{y}^{(i_2)},2h^2 \mathrm{I})  & &(\mathcal{T}_1)\\ 
  &&&\\
 &- \frac{2}{n\tilde{n}} \sum_{i=1}^{\tilde{n}} \sum_{j=1}^n \mathcal{N}(0;\tilde{y}^{(i)}-y^{(j)},2h^2 \mathrm{I})   && (\mathcal{T}_2)\\
 &&&\\
 &   -\lambda_1 \frac{1}{\tilde{\tilde{n}}} \sum_{k=1}^{\tilde{\tilde{n}}} \mathcal{N}(0;\tilde{y}^{(k)}-y^{(k)},2h^2 \mathrm{I})  &&(\mathcal{T}_3)\\
 &&&\\
  &  +\lambda_2 \ \mathcal{P}(\tilde{y}) && (\mathcal{T}_4)\\
  &&&\\
   &  +\lambda_3 \ \mathcal{P}(\phi) && (\mathcal{T}_5)\\
   &&&\\
    \end{array}
\end{equation}
with  $
\mathcal{N}(z;a,\Sigma)$
indicating a normal distribution for random vector $z$ with expectation $a$ and covariance matrix $\Sigma$. $\mathrm{I}$ is the identity matrix, $h$ is a user defined bandwidth and $\lambda_{1,2,3}$ are weights.
This paper aims at proposing an OT formulation for the terms 
$\mathcal{T}_2$ and $\mathcal{T}_3$  (see Sec. \ref{sec:OT})  as an alternative to $\mathcal{L}_2$  (presented in Sec. \ref{sec:L2}). In particular we show that these terms corresponds to robust Wasserstein distances  where the bandwidth $h$ (Eq. \ref{eq:1}) enables the seamless control of the level of robustness in a similar fashion as the scale parameter controlling  M-estimators \cite{B-Huber}. This reformulation allows the following contributions: first, to extend OT in supervised and semi-supervised scenarios, and second to propose  a robust Wasserstein cost (Sec. \ref{sec:OT}). We start first by explaining in more detail the notations used and the $\mathcal{L}_2$   cost function.

%%%%%%%%%%%%%%%%%%%%%%%%%%
%%%%%%%%%%%%%%%%%%%%%%%%%%
\section{$\mathcal{L}_2$ divergence}
\label{sec:L2}

\begin{figure}[!h]
\begin{center}
\includegraphics[width=.8\linewidth]{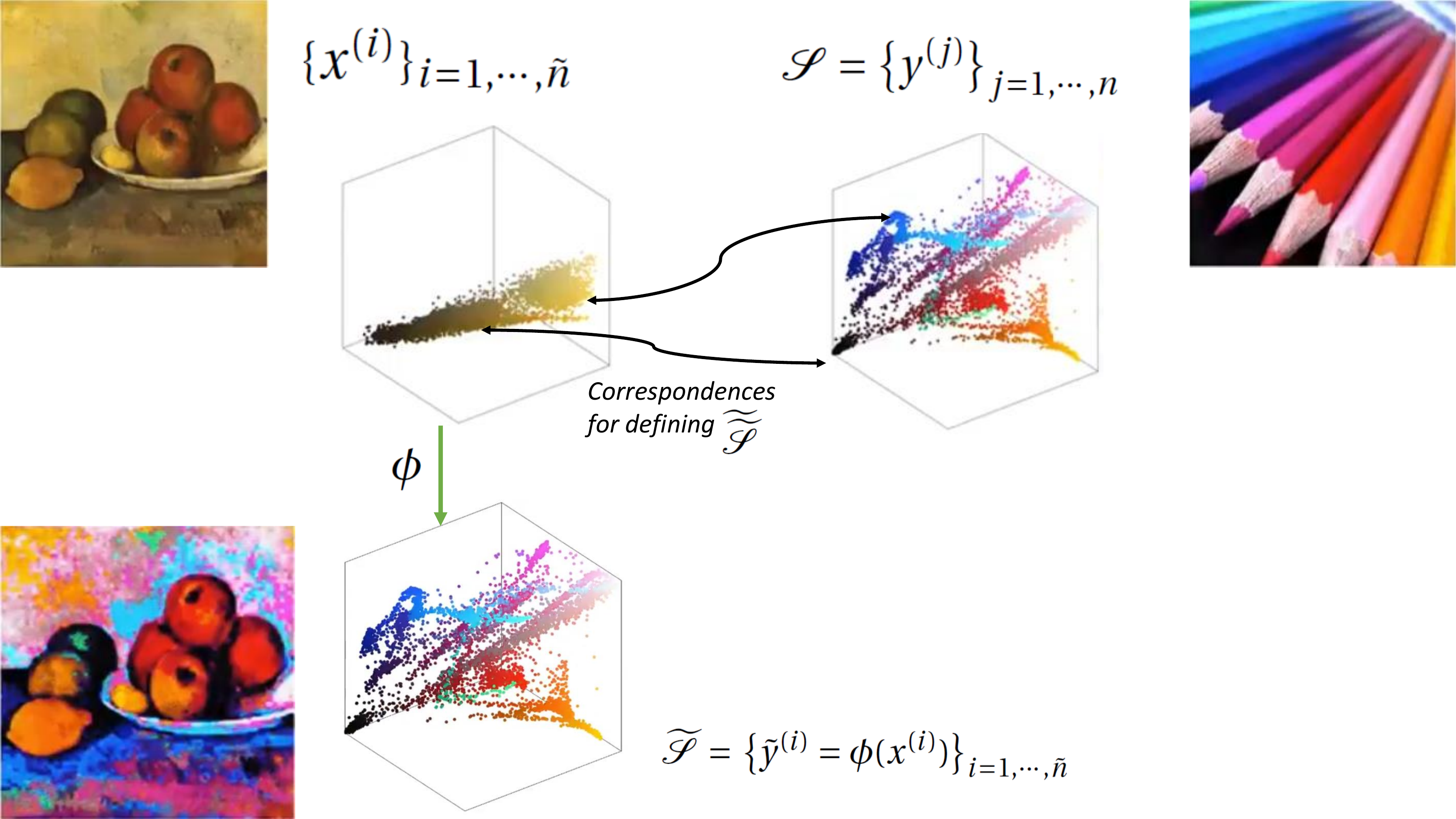}
\caption{Illustration of the  datasets considered in this paper. Data points correspond to triplets of values (e.g. RGB) for colour transfer. The aim of colour transfer is to estimate a mapping function $\phi$ to recolour the pixels of an image using the colour palette of another. When $\phi$ is well chosen, the point cloud  $\widetilde{\mathcal{S}}$ should overlap well with $\mathcal{S}$. }
\end{center}\label{fig:2}
\end{figure}

We consider that the following are available:
\begin{itemize}
    \item a dataset  $\mathcal{S}=\left \lbrace y^{(j)}\right\rbrace_{j=1,\cdots,n}$: the term $\mathcal{T}_0$ (Eq. \ref{eq:1}) uses the samples from this dataset.  
    \item a dataset  $\widetilde{\mathcal{S}}=\left\lbrace \tilde{y}^{(i)}=\phi(x^{(i)})\right\rbrace_{i=1,\cdots,\tilde{n}}$ computed using a transfer (or mapping) function $\phi$ on  data points $\lbrace x^{(i)}\rbrace _{i=1,\cdots,\tilde{n}}$. The term $\mathcal{T}_1$ (Eq. \ref{eq:1}) uses the samples from this dataset. 
    \item  a dataset of correspondences $\widetilde{\widetilde{\mathcal{S}}}=\left\lbrace ( y^{(k)},\tilde{y}^{(k)}=\phi(x^{(k)}))\right\rbrace_{k=1,\cdots,\tilde{\tilde{n}}}$ : the term $\mathcal{T}_3$ (Eq. \ref{eq:1}) uses the samples from this dataset. 
\end{itemize}
All data points have the same dimension (i.e.  $\dim (y^{(l_1)})=\dim(\tilde{y}^{(l_2)})$) for any samples taken from $\mathcal{S}$, $\widetilde{\mathcal{S}}$ or  $\widetilde{\widetilde{\mathcal{S}}}$.
Figure \ref{fig:2} shows an illustration of our datasets in the context of colour 
transfer\footnote{Images from the video posted at 
\url{https://twitter.com/gabrielpeyre/status/979605863295053826} have been used for designing Fig. \ref{fig:2}.}.
In this $\mathcal{L}_2$  framework \cite{GROGAN201939}, only one random vector (r.v.) $y$ is defined. Using $\mathcal{S}$ 
and $\widetilde{\mathcal{S}}$, two probability density functions noted $\mu(y)$ and $\tilde{\mu}(y|\phi)$ respectively are computed for r.v. $y$ as  kernel density estimates with a Normal kernel (or Gaussian Mixture Models): 
$$
\mu(y)= \frac{1}{n}\sum_{y^{(j)}\in \mathcal{S}} \mathcal{N}(y;y^{(j)},h^2)
$$ 
 and 
 $$
 \tilde{\mu}(y|\phi) = \frac{1}{\tilde{n}}\sum_{y^{(i)}\in \widetilde{\mathcal{S}}} \mathcal{N}(y;\tilde{y}^{(i)},h^2)
 $$
 The unknown mapping function $\phi$ transforms the samples in $\widetilde{\mathcal{S}}$  
that act  as the means of the normal kernels in the mixture $ \tilde{\mu}(y|\phi)$. Hence, $\tilde{\mu}$ can be warped onto $\mu$ by finding the appropriate function $\phi$. The best choice for  function $\phi$ can be chosen as minimising the Euclidean $\mathcal{L}_2$ distance  between $\mu$ and $\tilde{\mu}$   defined as \cite{JianPAMI2011}:
\begin{equation}
    \mathcal{L}_2 (\mu,\tilde{\mu}) =\| \mu-\tilde{\mu}\|^2=\int (\mu(y)-\tilde{\mu}(y|\phi))^2 \ dy = \underbrace{\| \mu\|^2}_{\mathcal{T}_0} \underbrace{- 2 \langle \mu | \tilde{\mu} \rangle}_{\mathcal{T}_2} + \underbrace{ \| \tilde{\mu} \|^2 }_{\mathcal{T}_1}
    \label{eq:L2}
\end{equation}
from which terms $\mathcal{T}_{0,1,2}$ in the cost function $\mathcal{C}(\phi)$ originate (Eq. \ref{eq:1}). 
Such a formulation of $\mathcal{L}_2$ has been used
    for colour transfer \cite{GROGAN201939} and shape registration \cite{JianPAMI2011,ARELLANO201612}. 
  The connection between $\mathcal{L}_2$  with robust M-estimators has also been   shown \cite{BasuBiometrika1998,ScottTechnometrics2001,JianPAMI2011}.
 
\paragraph{Removing  $\mathcal{T}_0$ from the cost function $\mathcal{C}$.} $\mathcal{T}_0$ does not depends on $\phi$ and can be discarded, shortening $\mathcal{L}_2$ into $\mathcal{L}_2E$ \cite{ScottTechnometrics2001} for estimating $\phi$. Both $\mathcal{T}_0$ and $\mathcal{T}_1$ correspond to entropies since $-\log\left(\|\mu\|^2\right)$ and $-\log\left(\|\tilde{\mu}\|^2\right)$ are the quadratic Renyi entropies of $\mu$ and $\tilde{\mu}$ respectively \cite{GROGAN201939}.

\paragraph{Using correspondences.} The term $\mathcal{T}_3$   to account for correspondences in $\widetilde{\widetilde{\mathcal{S}}}$, is explained  intuitively with notation $-2 \langle \mu | \tilde{\mu} \rangle$ by Grogan et al \cite{Grogan:2017:UII:3150165.3150171}, 
where this time $\mu$ and $\tilde{\mu}$ are likewise  kernel density estimates (with Normal kernel) using only observations in the dataset of correspondences $\widetilde{\widetilde{\mathcal{S}}}$:
$$
\tilde{\mu}(y)=\frac{1}{\tilde{\tilde{n}}} \sum_{y^{(k_1)}\in \tilde{\tilde{\mathcal{S}} }} \mathcal{N}(y;y^{(k_1)}, h^2\mathrm{I}) \quad \text{and} \quad \tilde{\mu}(y)=\frac{1}{\tilde{\tilde{n}}} \sum_{y^{(k_2)}\in \tilde{\tilde{\mathcal{S}} }} \mathcal{N}(y;\tilde{y}^{(k_2)}, \tilde{h}^2\mathrm{I}) 
$$
and the scalar product  $\langle \mu | \tilde{\mu} \rangle$ then corresponds to:
\begin{equation}
\langle \mu | \tilde{\mu} \rangle=\frac{1}{\tilde{\tilde{n}}^2}  \sum_{y^{(k_1)}\in \tilde{\tilde{\mathcal{S}} }} \sum_{y^{(k_2)}\in \tilde{\tilde{\mathcal{S}} }} \mathcal{N}(y^{(k_1)};\tilde{y}^{(k_2)}, (h^2+\tilde{h}^2)\mathrm{I}) 
\label{Eq:double:sum}
\end{equation}
Hence the notation $\langle \mu | \tilde{\mu} \rangle$  is  not  mathematically  correct  to explain  $\mathcal{T}_3$ (i.e note the single sum for $\mathcal{T}_3$ in Eq. \ref{eq:1} versus the double sum appearing in Eq. \ref{Eq:double:sum}). So
even if the intuition for $\mathcal{T}_3$  is sound and proves to be  efficient  in practice against  the state of the art techniques for colour transfer \cite{Grogan:2017:UII:3150165.3150171,GROGAN201939}, its origin cannot be explained mathematically with $\mathcal{L}_2$ and we provide next a better  explanation  for  $\mathcal{T}_3$  based on Optimal Transport.

%%%%%%%%%%%%%%%%%%%%%%%%%%%%%%%%%%%%%%%%%%%%%%%%%
%%%%%%%%%%%%%%%%%%%%%%%%%%%%%%%%%%%%%%%%%%%%%%%%%
\section{Optimal Transport}
\label{sec:OT}

We propose to reformulate both $\mathcal{T}_2$ and $\mathcal{T}_3$ from an OT perspective.
OT aims at choosing $\phi$ with the minimum transport (displacement) cost between two random vectors noted $y$  and $\tilde{y}$. The OT cost function is expressed here with the  Wasserstein distance \cite{MAL-073} as follow:
\begin{equation}
\mathcal{W}(\mu,\tilde{\mu})=\min_{\gamma} \left \lbrace \int \int c(y,\tilde{y}) \ \gamma(y,\tilde{y}) \ dy \ d\tilde{y}=\langle c| \gamma \rangle \right\rbrace
\label{eq:OT}
\end{equation}
where $c$ is a cost often chosen as $c(y,\tilde{y})=\|y-\tilde{y}\|^2 $ (quadratic Wasserstein distance), and $\gamma$ is the joint probability density function of  $y$  and $\tilde{y}$ having $\mu$ and $\tilde{\mu}$ for marginals respectively i.e. $\int \gamma(y,\tilde{y})\ dy=\tilde{\mu}(\tilde{y}) $ and $\int \gamma(y,\tilde{y}) \ d\tilde{y}=\mu(y)$.
We first present our choices for these distributions (Sec. \ref{sec:parametric:pdf}) and then propose a new robust cost 
(in Sec. \ref{sec:robust:cost}).
An alternative OT based explanation for terms $\mathcal{T}_2$ and $\mathcal{T}_3$ then emerges  (Sec. \ref{sec:new:T2:T3}).

%%%%%%%%%%%%%%%%%%%%%%%%%%%%%%%%%%%%%%%%%

\subsection{Models for $\gamma_{\phi}$, $\mu$ and $\tilde{\mu}_{\phi}$}
\label{sec:parametric:pdf}

Kernel density estimates with  Normal kernels are used as joint density functions  $\gamma_{\phi}$ and using the datasets available, three estimates of $\gamma_{\phi}\in\lbrace \gamma_u,\gamma_s,\gamma_{s+u}\rbrace$ can be proposed: 
\begin{itemize}
\item  using independent sets $\mathcal{S}$ and $\widetilde{\mathcal{S}} $ (unsupervised scenario i.e without  correspondences):
\begin{equation}
\gamma_u(y,\tilde{y}| \phi) =\left(\frac{1}{n} \sum_{y^{(j)}\in \mathcal{S}} \mathcal{N}(y;y^{(j)},h^2\mathrm{I}) \right) \times \left(\frac{1}{\tilde{n}} \sum_{\tilde{y}^{(i)}\in \widetilde{\mathcal{S}}} \mathcal{N}(\tilde{y};\tilde{y}^{(i)},\tilde{h}^2\mathrm{I})
\right) 
\end{equation} 
with  the marginals 
$$\mu_u(y)=\frac{1}{n} \sum_{y^{(j)}\in \mathcal{S}} \mathcal{N}(y;y^{(j)},h^2\mathrm{I}) $$
and  
$$
\tilde{\mu}_u(\tilde{y}|\phi)=\frac{1}{\tilde{n}}\sum_{\tilde{y}^{(i)}\in \widetilde{\mathcal{S}}} \mathcal{N}(\tilde{y};\tilde{y}^{(i)},\tilde{h}^2\mathrm{I})
$$

\item  using the set of correspondences $\widetilde{\widetilde{\mathcal{S}}}$ (supervised):
\begin{equation}
\gamma_s(y,\tilde{y}|\phi) =\frac{1}{\tilde{\tilde{n}}}\sum_{(y^{(k)},\tilde{y}^{(k)})\in \widetilde{\widetilde{\mathcal{S}}}} \mathcal{N}(y;y^{(k)},h^2\mathrm{I}) \ \mathcal{N}(\tilde{y};\tilde{y}^{(k)},\tilde{h}^2 \mathrm{I})
\end{equation}
providing  the  marginals 
$$
\mu_s(y)=\frac{1}{\tilde{\tilde{n}}} \sum_{y^{(k)}\in \widetilde{\widetilde{\mathcal{S}}}} 
\mathcal{N}(y;y^{(k)},h^2\mathrm{I})$$
and:  
$$
\tilde{\mu}_s(\tilde{y}|\phi)=\frac{1}{\tilde{\tilde{n}}}
\sum_{\tilde{y}^{(k)}\in \widetilde{\widetilde{\mathcal{S}}}}  \mathcal{N}(\tilde{y};\tilde{y}^{(k)},\tilde{h}^2\mathrm{I})
$$

\item Using all datasets, the following mixture can be considered (semi-supervised):
\begin{equation}
\gamma_{s+u}(y,\tilde{y}|\phi)= (1-\lambda) \ \gamma_u(y,\tilde{y}|\phi)+ \lambda \ \gamma_s(y,\tilde{y}|\phi)
\end{equation}
where $0\leq \lambda \leq 1$  is a parameter controlling the importance between the estimates $\gamma_u$ and $\gamma_s$. In this case, the marginals are:
$$
\mu_{s+u}(y)= (1-\lambda) \ \mu_u(y)+ \lambda \ \mu_s(y)
$$
and 
$$
\tilde{\mu}_{s+u}(\tilde{y}|\phi)= (1-\lambda) \ \tilde{\mu}_u(\tilde{y}|\phi)+ \lambda \ \tilde{\mu}_s(\tilde{y}|\phi)
$$
\end{itemize}
Note that these models noted $\gamma_{\phi}\in\lbrace \gamma_u,\gamma_s,\gamma_{s+u}\rbrace$ are parameterized by $\phi$ via the samples $\tilde{y}^{(l)}$ in $\widetilde{\mathcal{S}}$
and $\widetilde{\widetilde{\mathcal{S}}}$.
The bandwidths $h$ and $\tilde{h}$  are user defined and using $h=\tilde{h}=0$ enables the recovery of the empirical pdf estimates with Dirac kernels.

%%%%%%%%%%%%%%%%%%%%%%%%%%%%%%%%%%%
\subsection{Robust cost $c_G(y,\tilde{y})$}
\label{sec:robust:cost}

Concave functions $g$ to define costs $c$ of the  form $c(y,\tilde{y})=g(|y-\tilde{y}|)$  have  been suggested for robustness  \cite{DelonSIAM2012}. 
Here, we go further by proposing the following  robust cost:
\begin{equation}
c_G(y,\tilde{y})=A- \mathcal{N} (y;\tilde{y}, h_c^2 \mathrm{I})%\mathcal{N} (y;\tilde{y}, 2h^2 \mathrm{I})
\end{equation}
where $A$ is a constant that can be added if one need  to enforce a positive cost $c_G$. Our cost $c_G$ is convex near the origin $\|y-\tilde{y}\|\sim 0$ and then becomes concave as the difference $\|y-\tilde{y}\|$ increases.  We also note that: 
\begin{equation}
\langle c_G| \gamma \rangle =A - \int \int \mathcal{N} (y;\tilde{y}, h_c^2 \mathrm{I}) \ \gamma(y,\tilde{y}) \ dy \ d\tilde{y}
\end{equation}
since $\gamma$ integrates to 1 by definition.
In practice, for  estimation of  $\phi$ that minimizes this cost, the constant $A$ does not matter and can be set $A=0$.

%%%%%%%%%%%%%%%%%
\subsubsection{Relation to M-estimators}

\begin{figure}[!h]
\begin{center}
\begin{tabular}{cc}
\includegraphics[width=.45\linewidth]{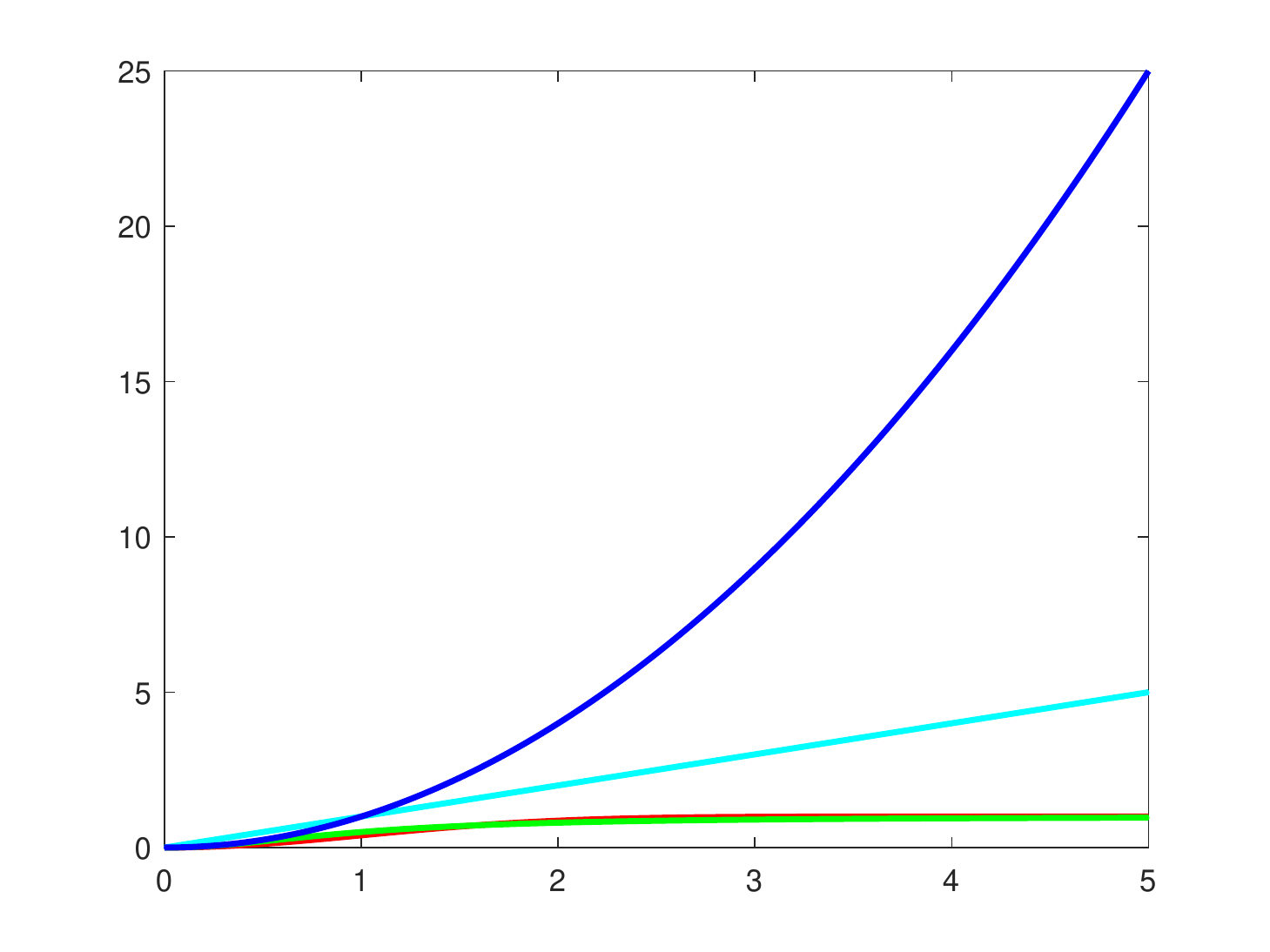}&
\includegraphics[width=.45\linewidth]{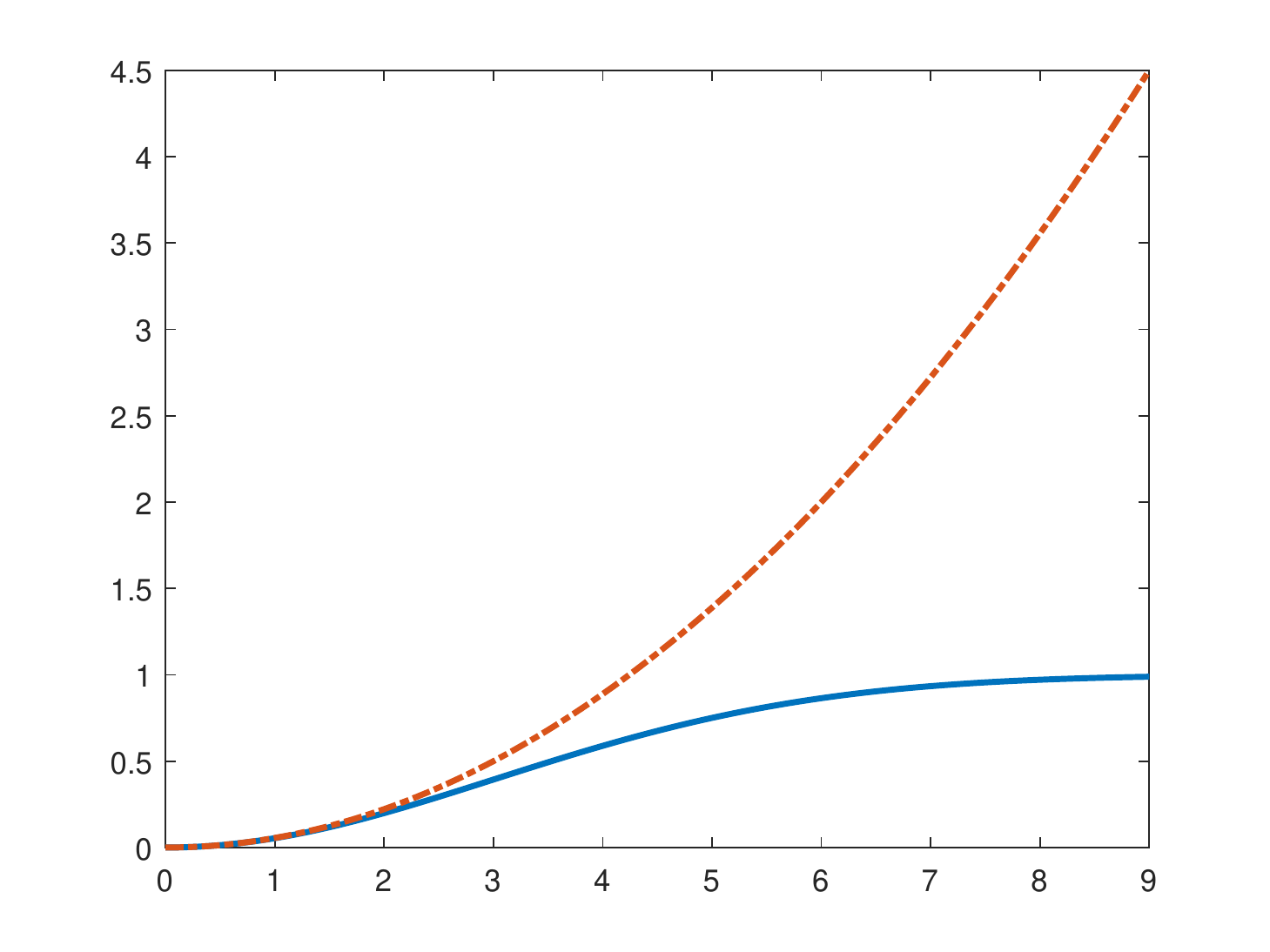}\\
(a)& (b)\\
\end{tabular}
\caption{Left (a): Functions  $\rho_{ls}(\epsilon)=\epsilon^2$ (blue), $\rho_{abs}(\epsilon)=|\epsilon|$ (cyan), Welsch-Leclerc $\rho_{G}(\epsilon)=1-\exp\left(-\epsilon^2/2\right)$  (red) and Geman-McClure $\rho_{GM}(\epsilon)=\frac{\epsilon^2}{\epsilon^2+1}$ (green). Right (b): comparison $\rho_G$ (solid blue line) with its Taylor approximation (orange dash dot)  with scale $\sigma=3$ for a range $\epsilon\in (0;3\sigma)$ - the approximation is good for  $\left | \frac{\epsilon}{\sigma}\right| << 1 $.}
\label{fig:rho}
\end{center}
\end{figure}

With the more familiar notation for error $\epsilon=\|y-\tilde{y}\|$, our robust cost $c_G$ is proportional to the Welsch-Leclerc loss $\rho_G$ \cite{DBLP:journals/corr/Barron17}:
\begin{equation}
\rho_G(\epsilon)=1-\exp\left(-\frac{1}{2} \left(\frac{\epsilon}{\sigma}\right)^2\right) \end{equation}
which is a well-known hard redescending M-estimating function with scale parameter $\sigma=h_c$ 
\cite{B-Huber,6152129,DAHYOT2013788,DBLP:journals/corr/Barron17}.
The more the chosen  function $\rho$ penalises large errors $\epsilon$, the more it is robust to outliers. See for instance in Fig. \ref{fig:rho}(a) how the hard redescending  functions $\rho_{GM}$ (for Geman-McClure loss \cite{Dahyot2004,DBLP:journals/corr/Barron17}) and $\rho_{G}$ have an upper finite limit (equal to 1) when $\epsilon \rightarrow +\infty$ and thus prevent high residuals (outliers) to overly contribute too much when estimating $\hat{\phi}$. The non-robust Least Square function $\rho_{LS}$ is also shown and corresponds here to the quadratic Wasserstein cost $c(y,\tilde{y})=\|y-\tilde{y}\|^2$ that is not robust to gross errors.

%%%%%%%%%%%%%%%%%
\subsubsection{Relation of robust cost $c_G$ to Wasserstein distance}

When the bandwidth $h_c$ (or scale parameter $\sigma$) is very  very large  compared to $\epsilon$, using Taylor approximation of the cost  shows that (cf. Fig. \ref{fig:rho}(b)):
\begin{equation}
\rho_G(\epsilon)=1-\exp\left(-\frac{\epsilon^2}{2\sigma^2}\right) \sim \frac{\epsilon^2}{2\sigma^2} \quad \text{for} \quad \epsilon<<\sigma
\end{equation}
making our cost $c_G$  proportional to the one used in  the quadratic Wasserstein distance. 
The bandwidth $h_c$  allows for the modulation of the cost from the non robust Euclidean  distance  ($h_c\rightarrow \infty$) to a more robust cost ($h_c$ small) for penalising high differences  $\|y-\tilde{y}\|$ (or outliers).

%%%%%%%%%
\subsection{OT perspective for terms $\mathcal{T}_2 $ and $\mathcal{T}_3 $}
\label{sec:new:T2:T3}

Using the definitions of our cost $c_G=- \mathcal{N} (y;\tilde{y}, h_c^2 \mathrm{I})$ and  our joint probability density functions  $\gamma_{\phi}\in\lbrace \gamma_u,\gamma_s,\gamma_{s+u}\rbrace$ (cf. Sec. \ref{sec:parametric:pdf}), we note that: 
\begin{equation}
    \langle c_G| \gamma_u \rangle = \frac{1}{n\tilde{n}} \sum_{i=1}^{\tilde{n}} \sum_{j=1}^n \mathcal{N}\left(0;\tilde{y}^{(i)}-y^{(j)},(h^2+\tilde{h}^2+h_c^2) \ \mathrm{I}\right)
\end{equation}
hence it is equivalent to the term 
 $\mathcal{T}_2$ (since the bandwidths are user defined).
  Likewise we note 
 \begin{equation}
     \langle c_G| \gamma_s \rangle=  \frac{1}{n_c} \sum_{k=1}^{n_c} \mathcal{N}\left(0;\tilde{y}^{(k)}-y^{(k)},(h^2+\tilde{h}^2+h_c^2) \ \mathrm{I}\right)
 \end{equation}
  which is equivalent to $\mathcal{T}_3 $ (Eq. \ref{eq:1}) introduced by Grogan et al to take advantage of correspondences \cite{GROGAN201939}. Since the weight $\lambda_{1}$ was chosen in an ad hoc fashion, we can  propose a more elegant alternative form combining     $\mathcal{T}_2 $ and $\mathcal{T}_3 $ into a new term $\mathcal{T}$ using the estimate $\gamma_{s+u}$:
\begin{equation}
   \langle c_G| \gamma_{s+u} \rangle= (1-\lambda) \ \langle c_G| \gamma_u \rangle+\lambda \ \langle c_G | \gamma_s \rangle \quad \quad (\mathcal{T})
\end{equation}
With the OT formulation (Eq. \ref{eq:OT}), Grogan et al's  estimation (terms $\mathcal{T}_2 $ and $\mathcal{T}_3 $, Eq. \ref{eq:1}) can be rewritten:
\begin{equation}
\hat{\phi} = \arg \min_{\phi} \left\lbrace \mathcal{W}(\mu_{s+u},\tilde{\mu}_{s+u})=\langle c_G| \gamma_{s+u} \rangle \right\rbrace
\end{equation}
to which entropic terms on the marginals $\mu $ and $\tilde{\mu}$ ($\mathcal{T}_0$ and $\mathcal{T}_1$) can be added along with other constraints on $\phi$ (e.g. $\mathcal{T}_4$ and $\mathcal{T}_5$). 
 
 When setting $h=\tilde{h}=0$ for simplicity (i.e. using empirical pdf estimates with Dirac kernels, Sec. \ref{sec:parametric:pdf}),  Grogan et al's  terms $\mathcal{T}_2$ and $\mathcal{T}_3$
are robust OT distances where the parameter $h_c$ in the robust cost $c_G$ controls the influence of outliers when performing estimation of the mapping function $\phi$ in the same way as the scale parameter for M-estimation.

\subsection{Parametric Modelling of the transfer function  $\phi$}

In practice, a parametric form of $\phi$ is used: Thin Plate Splines (TPS) have been used for colour transfer and shape registration \cite{JianPAMI2011,GROGAN201939,Grogan:2017:UII:3150165.3150171}. The term $\mathcal{T}_5$ in Eq. \ref{eq:1} corresponds to a smoothness constraint on the TPS solution \cite{JianPAMI2011,GROGAN201939}:
\begin{equation}
\mathcal{T}_5 =\lambda_3 \int \left\|\frac{\partial^2\phi(x)}{\partial^2 x}\right\|^2 dx 
\end{equation}
However TPS is not a convenient formulation when modelling transfer functions in  high dimensional spaces and Deep Neural Networks are now providing more powerful formulations for $\phi$.

\subsection{Interpretation and Generalization of the cost $c_G$} 

Our formulation of OT is equivalent to :
\begin{equation}
    \hat{\phi}=\arg \max_{\phi}  \int \int \mathcal{N}(y;\tilde{y},h_c^2 \mathrm{I})  \ \gamma_{\phi}(y,\tilde{y}) \ dy \ d\tilde{y} 
    \label{eq:myOT}
\end{equation}
where more generally $\mathcal{N}(y;\tilde{y},h_c^2 \mathrm{I})$ can be understood as a conditional pdf ($y$ given $\tilde{y}$ or vice versa since the Normal distribution is symmetric w.r.t. its mean). Using a flat prior for $\tilde{y}$ (e.g. $\tilde{y}\sim \mathcal{N}(\tilde{y};0, a\mathrm{I})$ with bandwidth $a$ very large to approximate a flat prior), then a model for the joint probability density function is available $\gamma_m(y,\tilde{y})=\mathcal{N}(y;\tilde{y},h_c^2 \mathrm{I})\times \mathcal{N}(\tilde{y};0, a\mathrm{I})$ and our OT formulation (Eq. \ref{eq:myOT}) is equivalent to:
\begin{equation}
    \hat{\phi}=\arg \max_{\phi}  \langle \gamma_m |\gamma_{\phi} \rangle  
\end{equation}
which has the same form as the cross product $ \langle \mu |\tilde{\mu} \rangle$ appearing in $\mathcal{L}_2$ (cf. Eq. \ref{eq:L2}): as indicated in \cite{GROGAN201939}, the main difference between the two frameworks lies in the modelling of one r.v. ($y$ in  $\mathcal{L}_2$, with notation $\langle .|.\rangle$ indicating integration over this one vector) or two r.v. ($y$ and $\tilde{y}$ in OT, $\langle .|.\rangle$ indicating integration over these two vectors). 
These scalar products between probability densities functions (joint, marginals or conditionals) are frequent  for robust estimation including for instance the Hough transform widely used in image processing \cite{4695834,DAHYOT2013788,Golden04}. 
While some robust costs can be identified as a negative log likelihood \cite{Dahyot2004,DBLP:journals/corr/Barron17}, we identify directly  our robust cost $c_G$ as a negative Multivariate Normal distribution instead.

%%%%%%%%%%%%%%%%%%%%%%%%%%%%%%%%%%%%%%%%%%%
\section{Final remarks}
\label{sec:conclusion}

We have proposed  a new generic formulation for Optimal Transport with the following advantages: 
\begin{itemize}
    \item it is robust:  our new robust cost $c_G(y,\tilde{y})=-\mathcal{N}(y;\tilde{y},h_c^2 \mathrm{I})$ is parameterised by a bandwidth $h_c$  that acts like the scale parameter of M-estimators. This bandwidth enables the control of the level of robustness and when  chosen very large, it makes our cost converge towards the standard (non robust) quadratic Wasserstein distance.
    \item Our formulation can seamlessly consider various scenarios e.g. unsupervised, supervised (with correspondences) or semi-supervised depending on the dataset(s) available. 
    \item Grogan et al  \cite{GROGAN201939}  propose the use of entropy terms for the marginals (e.g. $\tilde{\mu}$) that can be used in addition to (or instead of) an entropy on the joint pdf $\gamma$ \cite{MAL-073}.
    \item More generally, we have shown the commonality of these formulations ($\mathcal{L}_2$ and OT) in using scalar products between two p.d.fs. The main difference between $\mathcal{L}_2$ and OT is then  in the number of random vectors used in the formulation of this scalar product. We believe this thinking extends to the Gromov-Wasserstein formulation which defines 4 random vectors  \cite{Solomon:2016:EMA:2897824.2925903}.          
\end{itemize}
Beyond the impact of our formulation for colour transfer \cite{Grogan:2017:UII:3150165.3150171,GROGAN201939}, future work will investigate shape registration with correspondences (e.g.  for user interactions) and with  kernels other than Gaussian better suited to directional data \cite{GROGAN2018452}.

\section*{Acknowledgments}

This work is partly supported by a scholarship from Umm Al-Qura University, Saudi Arabia, and in part by a research grants from Science Foundation Ireland (SFI)  (Grant Number 15/RP/2776), and the ADAPT Centre for Digital Content Technology (www.adaptcentre.ie) that is funded under the SFI Research Centres Programme (Grant 13/RC/2106) and is co-funded under the European Regional Development Fund.

%%%%%%%%%%%%%%%%%%%%%%%%
%\appendix

%\bibliographystyle{apalike}
\bibliographystyle{plain}

\bibliography{References}

\end{document}